\documentclass[letterpaper]{article} 
\usepackage[preprint]{aaai2027}
\usepackage[hyphens]{url}  
\usepackage{graphicx} 
\usepackage{natbib}  
\usepackage{caption} 
\usepackage{algorithm}
\usepackage{algorithmic}

\usepackage{booktabs}
\usepackage{multirow}
\usepackage{amsmath,amssymb,amsfonts}
\usepackage{amsthm}
\usepackage{dsfont}
\newcommand{\clamp}{\operatorname{clamp}}
\newcommand{\ind}{\mathds{1}}
\newcommand{\Cov}{\operatorname{Cov}}
\newcommand{\Real}{\mathbb{R}}
\newcommand{\Ex}{\mathbb{E}}
\newcommand{\dgm}{\psi}
\newcommand{\logit}{\operatorname{logit}}
\DeclareMathOperator*{\argmax}{arg\,max}

\title{TRACER: Per-Tool Context Retention for LLM Agents via Consequence-Attributed Reinforcement Learning}

\author{
    Ziqi Lin,
    Ye Wu,
    Mengying Yang,
    Xu Liu,
    Yizhou Liu,
    Qiang Ke,
    Qin Guo
}
\affiliations{
}
\begin{document}
\maketitle

\begin{abstract}
Enterprise data agents answer business queries by chaining many tool calls over multiple reasoning steps, routinely accumulating hundreds
  of thousands of context tokens per session. Existing compression strategies typically allocate retention budgets without accounting
  for the downstream consequences of removing individual tool outputs. Aggressive compression may therefore trigger costly tool re-%
  invocations that offset the initial savings. We call this the \textbf{compression--consequence gap}. To close it, we propose \textbf{TRACER}, which formulates compression as a sequential per-tool decision problem. A lightweight REINFORCE policy
  assigns query-conditioned retention ratios using only information available at each compression event. Its consequence-aware
  objective jointly accounts for task success, total token consumption, and post-compression tool re-invocations. To improve credit
  assignment, TRACER uses a learned outcome model to compare the predicted consequences of the selected retention ratio with those of
  fully retaining each tool output. On held-out production queries across three compressor backends, TRACER reduces total token
  consumption by \textbf{29--46\%} relative to keeping all context while maintaining comparable or higher task success. Compared with a
  tool-type-conditional static policy, TRACER provides an additional \textbf{15--18\%} of token savings.
  Interventional rollouts show that the learned per-tool credit scores correlate with measured single-tool consequences. The learned
  policy also yields positive savings when transferred across agent backbones and compressor architectures, and reduces token
  consumption by \textbf{18--25\%} on five held-out LOCA-bench environments. These results demonstrate the value of consequence-aware,
  per-tool context retention for improving the efficiency of long-horizon language agents.
\end{abstract}

\section{Introduction}
\label{sec:intro}

Enterprise data agents answer business questions by chaining tool calls over many reasoning steps~\citep{yao2023react,schick2024toolformer}. Because each tool output is appended to the context and never discarded, the working memory grows monotonically. A single SQL result alone can exceed ten thousand tokens, and a typical task spans five to ten such calls. The accumulated context therefore routinely reaches hundreds of thousands of tokens, exceeding model windows, exhausting cost budgets, and degrading reasoning through diluted attention. Therefore, context compression becomes essential.

\begin{figure}[t]
    \centering
    \includegraphics[width=0.7\columnwidth]{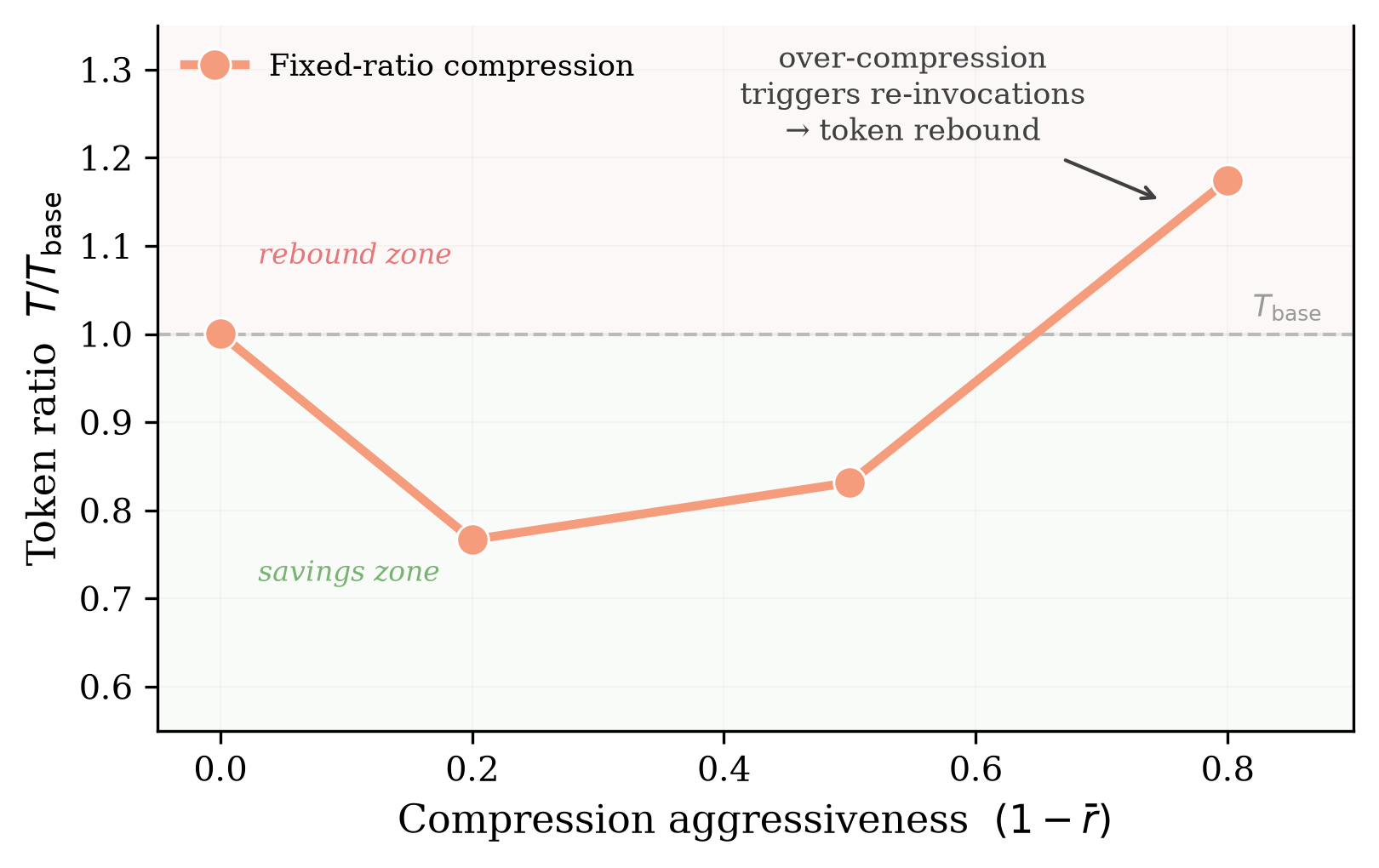}
    \caption{The compression--consequence gap. Token cost falls as compression becomes more aggressive, but total cost also accounts for downstream re-invocations.}
    \label{fig:compensation}
\end{figure}

Existing compression methods operate at various granularities: token-level pruning~\citep{jiang2023llmlingua,jiang2024longllmlingua,pan2024llmlingua2}, turn-level summarization~\citep{xu2024recomp}, and step-level dropping~\citep{agora2025}. Yet they all apply a uniform strategy across heterogeneous content. In an agentic session, tool outputs vary widely in downstream importance: a schema lookup may be referenced by every subsequent step, whereas a permission check is never revisited. What is needed is a query-conditioned, per-tool retention policy that allocates the token budget according to each output's future utility.

A deeper limitation is that existing methods evaluate compression in isolation. In an agentic loop, the model continues to act after compression. When a needed detail has been removed, the agent re-invokes the tool to recover it. For data agents, a single compensatory SQL call can consume tens of thousands of additional tokens. Because compression recurs as new outputs accumulate, such costs compound across steps. We call this mismatch between compression-time savings and downstream cost the \textbf{compression--consequence gap}, illustrated in Figure~\ref{fig:compensation}.

To close this gap we propose \textbf{TRACER}, a reinforcement-learning framework that dynamically determines the retention ratio for each tool output. The policy emits a continuous action per output, letting the token budget flow to outputs whose removal would trigger expensive recovery. The training reward combines task success, token efficiency, and the fraction of tools re-invoked after compression, so that savings purchased through costly recovery never yield a net gain. We train the policy with REINFORCE~\citep{williams1992simple} on full agent episodes.

A further challenge is credit assignment: each episode yields only one scalar reward across many per-tool decisions. To produce dense gradient signals, we introduce a learned outcome model that predicts per-tool consequences through single-tool counterfactuals.

  Our contributions are:
  \begin{itemize}
      \item We formalize the compression--consequence gap as a sequential, per-tool decision problem whose reward explicitly prices downstream re-invocations.
      \item We propose a reinforcement learning framework that dynamically determines per-tool retention ratios, parameterized by a Beta distribution over a bounded action space.
      \item We introduce a learned attributor that predicts per-tool re-invocation consequences through single-tool counterfactuals. On semantic compressors, the attributor reduces both token consumption and re-invocation rate. On positional compressors, it protects task success by steering retention toward high-consequence outputs.
  \end{itemize}

\section{Related Work}
\label{sec:related}

\paragraph{Context compression}
Context compression operates at multiple granularities, ranging from individual tokens to entire interaction steps.
  At the token level, methods prune by perplexity~\citep{jiang2023llmlingua,jiang2024longllmlingua} or distilled
  classifiers~\citep{pan2024llmlingua2}, and Selective Context drops low-self-information units~\citep{li2023selective}.
  At a coarser level, AGORA retains or discards entire observation--action pairs via counterfactual supervision~\citep{agora2025}.
Several recent methods learn when or how to compress the agentic context.MemGPT pages information across tiered storage~\citep{packer2023memgpt}; ACON, Focus, and Sculptor select a compressor mode per turn~\citep{yang2026acon,focus2026,sculptor2026}; SUPO co-trains summarization and action policies end-to-end~\citep{supo2026}; AdaCoM trains an external LLM via RL to manage context for a frozen agent~\citep{adacom2026}; and LLM-DCP formulates compression as an MDP over token removal~\citep{llmdcp2025}.
  These methods differ from TRACER primarily in decision granularity and
  action space. TRACER learns continuous retention ratios for individual
  tool outputs while keeping the underlying agent and compressor fixed.
  This design can potentially complement context managers that select
  higher-level compression operations, although both approaches optimize
  how information is retained in the agent context.

\paragraph{RL for language agents and credit assignment}
Prior RL methods for language agents assign credit at the step or turn level~\citep{wang2026steppo,li2025turnppo,wang2026ecpo}, exploit episode structure via graph decomposition or process rewards~\citep{cheng2026graphgpo,lu2026hisr,su2025eapo,cm2-2026,yuan2026vpr,wang2026rethinking}, or penalize redundant tool calls~\citep{toolaware2026,learnwhennot2026}.
All optimize the agent's action policy.
Our setting is the converse: we freeze the action policy and train only the context policy, using excess tool calls as a sparse credit signal.

To densify this signal, we adopt counterfactual credit assignment~\citep{foerster2018counterfactual,wolpert2001optimal,arjona2019rudder,chen2025emai}, which isolates one decision's contribution by comparing realized and alternative outcomes.
We instantiate this at the level of individual tool outputs and learn an outcome model that predicts the consequence of dropping each one.


\begin{figure*}[!t]
    \centering
    \includegraphics[width=0.9\textwidth]{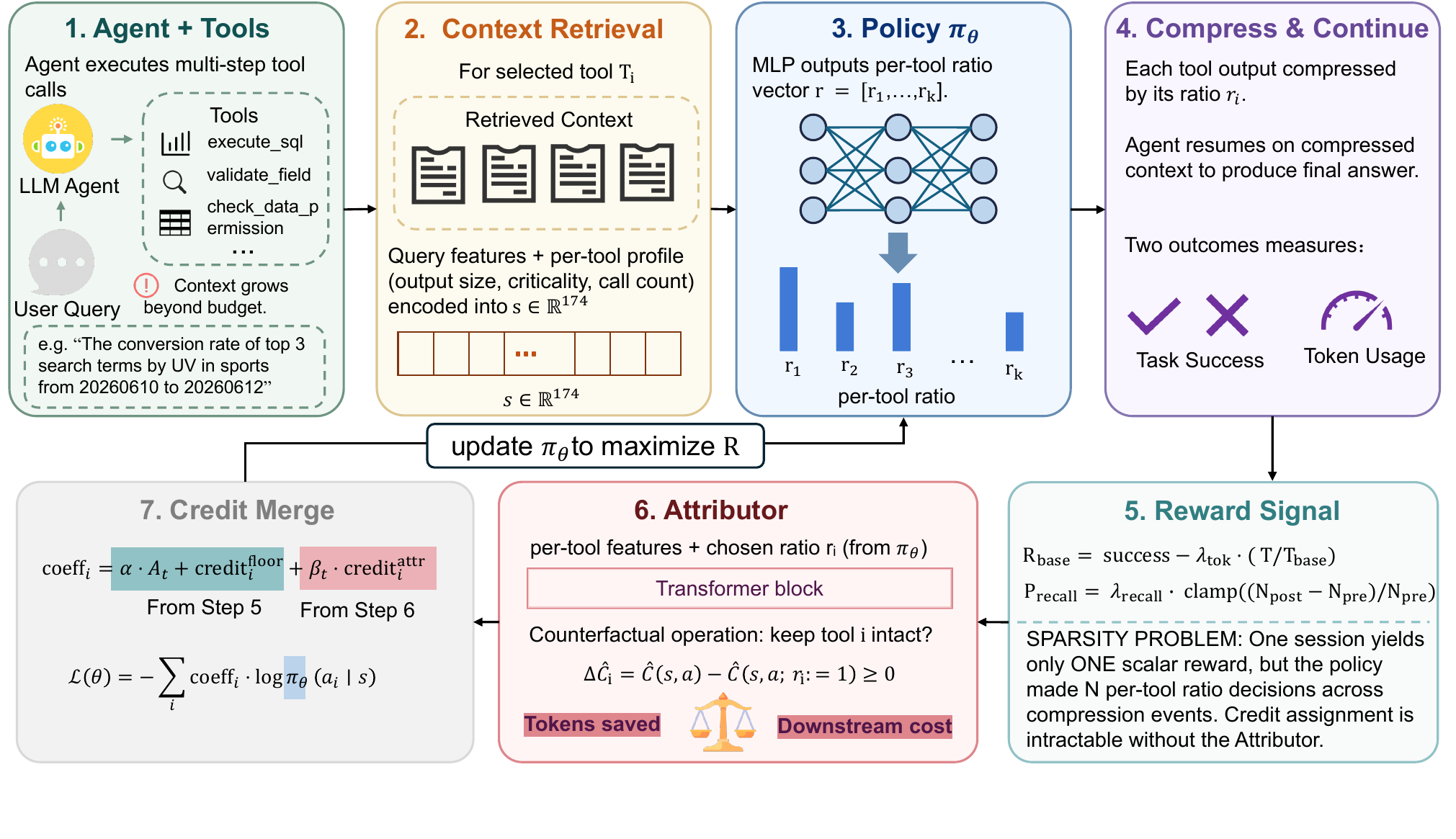}
    \caption{Overview of TRACER. TRACER uses a policy network to decide how much of each tool's information to keep during compression, trains that policy online from trajectory outcomes, and uses a post-hoc counterfactual model to attribute credit precisely to each tool.}
    \label{fig:framework}
\end{figure*}

\section{Method}
  \label{sec:method}

  As illustrated in Figure~\ref{fig:framework}, TRACER recasts context compression as a sequential decision problem.
  When a compression event occurs, the policy $\pi_\theta$ reads per-tool features and emits a heterogeneous retention ratio for each tool. A fixed compressor applies these ratios, and the agent continues reasoning over the compressed context. Throughout the trajectory we record task success, token usage, and tool re-invocations triggered after compression. These signals jointly drive an online REINFORCE update. Once the session ends, an outcome model attributes the observed consequences back to individual tools through single-tool counterfactuals, yielding a dense per-tool signal. We build TRACER through the following three core steps:

  \subsection{Problem Formulation}
  \label{sec:formulation}

The compression policy operates inside the agent's execution loop, so we first describe the agent and how its context grows, then define compression over this process.

The agent $\mathcal{A}$ runs in discrete steps. At step $t$, it reads the current context $c_t$, issues a tool
call, and appends the returned output to $c_t$. A session terminates when $\mathcal{A}$ emits a final answer or
reaches the iteration limit, and counts as successful when the answer matches the ground truth, within a 5\%
tolerance for numeric queries.

The context $c_t$ comprises the system prompt, the tool definitions, the dialogue history, and the accumulated
tool outputs. The last component dominates growth over a session. Once $|c_t|$ exceeds the token budget $\tau$,
a compression event shortens $c_t$ before the agent continues. Because the prompt and tool definitions are incompressible yet keep accumulating, $\tau$ is crossed more than once: a single session triggers $T_c$ compression events,
typically several even under a generous budget.

We cast this sequence of events as a sequential decision problem. Each session forms an episode, and each compression event $t = 1, \dots, T_c$ is a decision point. The state $s_t$ encodes the composition of $c_t$ through its segment shares
and per-tool footprints, including the tool-output share $f_{\text{tool}}$ that bounds the budget compression can
reclaim. The action $a_t$ assigns a retention ratio to each of the $N$ tool outputs present at the
event. A ratio chosen at event $t$ carries into later events: discarding an output that a subsequent step needs
forces a re-invocation whose cost falls on future decision points. This coupling
across events defines a sequential process, specified in full in Appendix~\ref{app:semimdp}.

  \subsection{Consequence-Aware Objective}
  \label{sec:reward}

  Minimizing token consumption alone is insufficient because it overlooks the re-invocation cost. Whenever compression discards information that is later needed, the agent must re-invoke tools to recover it. We therefore design a reward that decomposes into a session-level base reward and an event-local recall penalty.

  \paragraph{Session-level base reward}
  The first component combines task success with token efficiency and is measured only at episode end:
  \begin{equation}
  R_{\text{base},k} =
  \ind[k{=}T_c]\Big(\text{success}-\lambda_{\text{tok}}\,\clamp\big(\tfrac{T}{T_{\text{base}}},0.2,2\big)\Big),
  \label{eq:rbase}
  \end{equation}
  where $\text{success}\in\{0,1\}$ indicates whether the agent answered correctly, $T$ is the total session token count, and $T_{\text{base}}$ is the token total under the same query with no compression, averaged over five reference runs.

  \paragraph{Event-local recall penalty}
  While the base reward captures end-of-session outcomes, it cannot pinpoint which compression event caused a re-invocation.
  To provide a denser training signal, we introduce a penalty assessed immediately after each compression event:
  \begin{equation}
  P_{\text{recall},t} =
  \lambda_{\text{recall}}\,\clamp\Big(\tfrac{N_{\text{post},t}-N_{\text{pre},t}}{N_{\text{pre},t}},0,2\Big)\ge 0,
  \label{eq:precall}
  \end{equation}
  where $N_{\text{post},t}$ and $N_{\text{pre},t}$ denote the post-compression and reference tool-call counts within the same window. The reference counts are measured rather than assumed: we replay the same compression-trigger boundaries over the five keep-all timelines collected for that query, match windows by event order, and take $N_{\text{pre},t}$ to be their mean call count.

  \paragraph{Combined objective}
  Combining the two components yields the full objective:
  \begin{equation}
  J(\theta)=\Ex_{\tau\sim\pi_\theta}\big[R_{\text{base}}-P_{\text{recall}}\big],\quad P_{\text{recall}}=\textstyle\sum_t
  P_{\text{recall},t}.
  \label{eq:objective}
  \end{equation}
  The measured recall signal carries noise from natural call-count variation between rollouts. Appendix~\ref{app:prov} describes a provenance upgrade that removes this noise through causal matching.

  \subsection{Per-Tool Compression Policy}
  \label{sec:policy}

  \paragraph{Action parameterization}
  The retention ratio $a_i$ lives on a bounded interval. We use a Beta distribution whose support matches this interval by construction, avoiding truncation artifacts:
  \begin{equation}
  \tilde a_i\sim\mathrm{Beta}(\tilde m_i\kappa_0,\,(1-\tilde m_i)\kappa_0),
  \label{eq:beta}
  \end{equation}
  where $\tilde m_i=\sigma(\text{net}_i)$ is the network-predicted mean and $\kappa_0$ is a fixed concentration.
  At evaluation we replace sampling with the deterministic mean $\tilde m_i$.

  \paragraph{Network architecture}
  The policy is a two-layer MLP mapping state $s\in\Real^{174}$ to per-tool means $m\in\Real^N$ with $N$ tool slots. The slots are ordered positions for the individual tool-output instances present at the current compression event, not one position per canonical tool type: repeated calls to the same tool keep distinct output identifiers and therefore receive their own slot, their own retention ratio, and their own gradient. Each slot encodes query relevance, downstream reuse, redundancy, and recency. A presence mask ensures that only tools present at the current event receive gradients. Full feature definitions appear in Appendix~\ref{app:features}.

  \paragraph{Training}
  We maximize Eq.~\ref{eq:objective} with REINFORCE. A per-query EMA baseline $b_q$ normalized by per-query standard deviation handles difficulty variation. The per-tool policy loss is:
  \begin{equation}
  \mathcal{L}(\theta)=-\!\!\sum_{i\in\text{active}}\!\!\text{coeff}_i\cdot\log\pi_\theta(a_i\mid s),
  \label{eq:loss}
  \end{equation}
  where $\text{coeff}_i$ combines the session-level advantage with per-tool credit signals, defined next.

  The Beta parameterization makes this baseline choice principled rather than incidental. Writing $s_i=\partial\log\pi/\partial\tilde m_i$ for the score of slot $i$, the Beta log-moment identity gives $\Ex[s_i]=0$ (Appendix~\ref{app:beta}). Any constant or action-independent baseline is therefore admissible without introducing bias into this score, and the per-slot gradient takes the covariance form $g_i=\Ex[\text{credit}_i\,s_i]=\Cov(\text{credit}_i,s_i)$: a slot moves toward higher retention precisely when sampling a larger ratio co-varies with higher credit.

  \subsection{Counterfactual Credit Assignment}
  \label{sec:credit}

  The objective in Eq.~\ref{eq:objective} yields one scalar reward per session across many per-tool decisions. Standard REINFORCE assigns the same gradient magnitude to all tools, providing no signal about which tool caused a failure. We address this with a two-channel scheme. The first channel distributes the penalty uniformly as a model-free floor. The second adds per-tool directionality through a learned outcome model.

  \paragraph{Channel I: uniform penalty floor}
  The first channel distributes $P_{\text{recall},t}$ uniformly over the $k$ active tools:
  \begin{equation}
  \begin{aligned}
  \text{credit}^{\text{floor}}_i
    &=-P_{\text{recall},t}\,\frac{\ind[i\in\text{active}]}{k},\\
  \sum_i\text{credit}^{\text{floor}}_i
    &=-P_{\text{recall},t}.
  \end{aligned}
  \label{eq:floor}
  \end{equation}
  Even if the attribution channel degrades entirely, this floor guarantees a correct, if coarse, penalty signal.

  \paragraph{Channel II: counterfactual attribution via outcome model}
  Let $K$ denote the number of tools. We train an outcome model $f_\phi$, a 2-layer 4-head Transformer over $K$ tool tokens and one query token. It predicts per-tool re-invocation count $\hat n_i$, log token total $\hat T_{\log}$, and success probability $\hat p$. Self-attention captures cross-tool coupling: compressing one tool can force re-invocation of another.

  This model enables single-tool counterfactuals. For each tool $i$, we compare the predicted cost under the chosen ratio versus $r_i{:=}1$, holding all else fixed.
  Defining $\hat C(s,a)=\lambda_N\sum_j\hat n_j+\lambda_T\cdot2\tanh(\frac{1}{2}\frac{\exp\hat
  T_{\log}}{T_{\text{base}}})$, the per-tool credit becomes:
  \begin{equation}
  \text{credit}^{\text{attr}}_i=\underbrace{\lambda_{\text{save}}(1{-}r_i)\tfrac{\text{tok}_i}{T_{\text{base}}}}_{\text{tokens saved}}-\underbrace{\big(\hat C(s,a)-\hat C(s,a;\,r_i{:=}1)\big)}_{\text{downstream cost }\Delta\hat C_i\ge0}.
  \label{eq:attr}
  \end{equation}
  This signal is bidirectional: when savings exceed downstream cost, the signal encourages stronger compression; when the reverse holds, it discourages compression.

  \paragraph{Channel fusion}
  We fuse both channels into the per-tool gradient coefficient:
  \begin{equation}
  \text{coeff}_{i,t}=\underbrace{\alpha A_t+\text{credit}^{\text{floor}}_i}_{\text{measured}}+\underbrace{\beta_t\,\text{credit}^{\text{attr}}_i}_{\text{attributed}},
  \label{eq:coeff}
  \end{equation}
  where $\alpha$ scales the session-level advantage relative to the per-tool signals, and $\beta_t$ controls trust in $f_\phi$. We use a 20-episode warmup with $\beta{=}0$, followed by a 10-episode linear annealing to $\beta{=}1$, refreshing $f_\phi$ every 5 episodes thereafter. On held-out episodes, $f_\phi$ achieves Spearman $\rho{=}0.72$ for per-tool re-invocation prediction and Brier score $0.07$ for success prediction, confirming directional reliability. Algorithm~\ref{alg:tracer-main} summarizes the resulting training loop; the complete event-level algorithm appears in Appendix~\ref{app:algorithm}.

  \begin{algorithm}[t]
  \caption{TRACER training loop (condensed)}
  \label{alg:tracer-main}
  \begin{algorithmic}[1]
  \REQUIRE Queries $\mathcal{Q}$, compressor $\mathcal{C}$, agent $\mathcal{A}$, token budget $\tau$
  \STATE \textbf{Phase 0:} run $\mathcal{A}$ five times per query with $r_i{=}1$; store $T_{\text{base}}(q)$ and the reference timelines
  \FOR{episode $e=1,\dots,E$}
    \STATE $\beta_e\gets0$ if $e\le20$, else $\min\{1,(e{-}20)/10\}$
    \FOR{each compression event $t$ of the rollout}
      \STATE Build output-instance slots, state $s_t$, and the active mask
      \STATE Sample $a_{t,i}$ from the Beta policy (Eq.~\ref{eq:beta}); apply $\mathcal{C}$
      \STATE Match $N_{\text{pre},t}$ against the reference timelines; set $P_{\text{recall},t}$ (Eq.~\ref{eq:precall})
    \ENDFOR
    \STATE Observe success and session tokens $T$; form $R_{\text{base}}$ (Eq.~\ref{eq:rbase}) and the advantage $A$ against $b_q$
    \FOR{each recorded event $t$}
      \STATE $\text{credit}^{\text{floor}}$ (Eq.~\ref{eq:floor}); if $\beta_e{>}0$ also $\text{credit}^{\text{attr}}$ (Eq.~\ref{eq:attr})
      \STATE $\text{coeff}_{t,i}$ (Eq.~\ref{eq:coeff})
    \ENDFOR
    \STATE Update $\theta$ on the loss of Eq.~\ref{eq:loss}
    \IF{$e=20$, or $e>20$ and $(e{-}20)\bmod5=0$}
      \STATE Refit the outcome model $f_\phi$ from the replay buffer
    \ENDIF
  \ENDFOR
  \RETURN Frozen $\pi_{\theta^*}$; evaluate with its deterministic mean
  \end{algorithmic}
  \end{algorithm}

  \begin{table*}[t]
  \centering
  {\small
  \begin{tabular*}{\textwidth}{@{\extracolsep{\fill}}llcccccc@{}}
    \toprule
    \textbf{Compressor} & \textbf{Strategy} & \textbf{Success$\uparrow$} & \textbf{Tok.\ ratio$\downarrow$} & \textbf{Save} &
    \textbf{$p_\text{recall}$$\downarrow$} & \textbf{$\bar{N}_\text{tools}$} & \textbf{Cost/succ.$\downarrow$} \\
    \midrule
    \multirow{5}{*}{Truncation}
      & Uniform-0.5    & 0.73 & 1.026 & $-$3\% & 0.082 & 9.3 & 1.405 \\
      & Recency        & 0.77 & 1.086 & $-$9\% & 0.071 & 9.0 & 1.410 \\
      & Token-prop.    & 0.69 & 1.112 & $-$11\% & 0.091 & 9.5 & 1.612 \\
      & Tool-type-cond. & 0.76 & 0.841 & 16\% & 0.045 & 8.5 & 1.107 \\
      & \textbf{TRACER} & \textbf{0.77} & \textbf{0.694} & \textbf{31\%} & 0.021 & 7.8 & \textbf{0.901} \\
    \midrule
    \multirow{5}{*}{Summarization}
      & Uniform-0.5    & 0.80 & 0.832 & 17\% & 0.105 & 9.6 & 1.040 \\
      & Recency        & 0.80 & 1.194 & $-$19\% & 0.088 & 9.1 & 1.493 \\
      & Token-prop.    & 0.77 & 1.231 & $-$23\% & 0.098 & 9.4 & 1.599 \\
      & Tool-type-cond. & 0.79 & 0.723 & 28\% & 0.052 & 8.8 & 0.915 \\
      & \textbf{TRACER}  & \textbf{0.81} & \textbf{0.541} & \textbf{46\%} & 0.015 & 7.2 & \textbf{0.668} \\
    \midrule
    \multirow{5}{*}{Self-Info}
      & Uniform-0.5    & 0.62 & 1.207 & $-$21\% & 0.128 & 10.2 & 1.947 \\
      & Recency        & 0.71 & 1.242 & $-$24\% & 0.112 & 10.4 & 1.749 \\
      & Token-prop.    & 0.66 & 1.187 & $-$19\% & 0.121 & 10.0 & 1.798 \\
      & Tool-type-cond. & 0.70 & 0.891 & 11\% & 0.062 & 9.1 & 1.273 \\
      & \textbf{TRACER}  & \textbf{0.75} & \textbf{0.706} & \textbf{29\%} & 0.025 & 8.1 & \textbf{0.941} \\
    \bottomrule
    \end{tabular*}
  }
  \caption{Main comparison: TRACER vs.\ fixed and heuristic strategies.
  Token ratio = $T/T_\text{keep-all}$. Cost/succ.\ = Tok.\ ratio $/$ Success.}
  \label{tab:main}
  \end{table*}

\begin{figure*}[t]
    \centering
    \includegraphics[width=\textwidth]{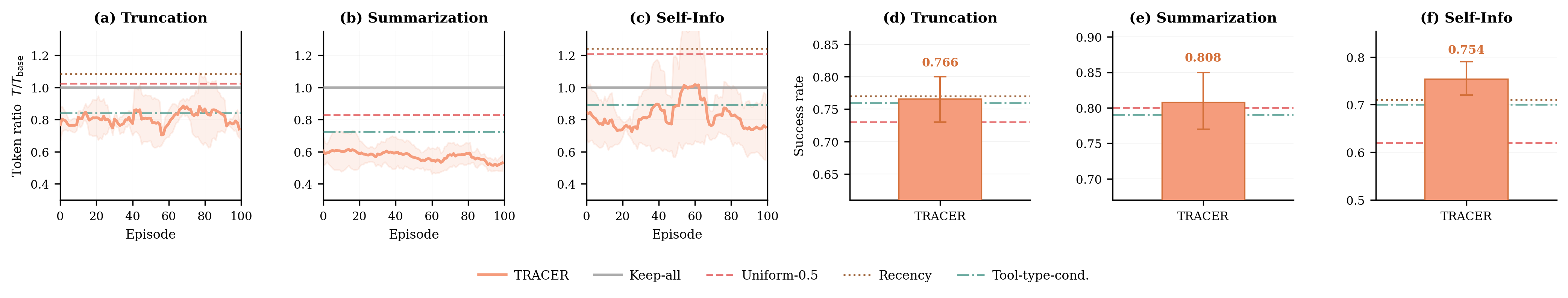}
    \caption{TRACER vs.\ fixed-strategy baselines across three compressors. Panels (a)--(c): token-ratio training curves (5 seeds, 15-episode rolling window, $\pm$1 std). Panels (d)--(f): task success rate with 95\% CI.}
    \label{fig:rl_vs_fixed}
\end{figure*}

\section{Experiments}
\label{sec:experiments}

\subsection{Setup}
\label{sec:setup}

\paragraph{Platform}
We evaluate on a production data-analysis agent deployed at a large e-commerce company. The agent answers business questions using tools for knowledge search, SQL execution, permission checks, table lookup, and date resolution. The agent backbone is Claude Sonnet 4.6~\citep{anthropic2025claude4} with a 200K-token context window, held fixed throughout all experiments.

Because our policy relies on tool-level features rather than model internals, it does not depend on any particular backbone. We evaluate three compressor backends ordered by model reliance. Uniform truncation needs no model. Self-Info, a clause-level variant of Selective Context~\citep{li2023selective}, ranks clauses by self-information scored by Qwen-2.5-0.5B~\citep{qwen2.5}. Summarization uses Qwen-3.7-Max~\citep{qwen3} to condense text semantically. A fourth AGORA-style~\citep{agora2025} step-level compressor is held out to test cross-compressor transfer.

\paragraph{Queries}
We collect 120 queries from production logs spanning three difficulty tiers: 25 simple lookups, 55 multi-step queries combining schema retrieval with SQL, and 40 multi-table permission-gated joins. We split them into 80 training and 40 held-out queries, stratified by difficulty, each with a human-verified answer. Every RL configuration trains for 100 episodes per seed with 5 seeds. Held-out evaluation uses frozen deterministic policies.

\paragraph{Baselines}
All strategies share the same compressor backend. Keep-all serves as the success ceiling and token baseline. Uniform-0.5 applies a constant ratio. Recency decays the ratio linearly with age, $r_i=0.1+0.8(\text{pos}_i-1)/(N-1)$, where $\text{pos}_i$ indexes invocation order, so the most recent output retains $0.9$ and the oldest $0.1$. Token-proportional penalizes size, $r_i=1-0.7\,\text{size}_i/\max_j\text{size}_j$, so the largest output retains $0.3$. Tool-type-conditional assigns a fixed ratio per canonical tool type, tuned by greedy coordinate descent over $\{0.2,0.3,\dots,0.9\}$ visiting types in decreasing mean-output-size order for two passes, each coordinate maximizing success rate minus $0.3$ times token ratio on the training queries. It isolates whether tool identity alone suffices without query-conditioned adaptation.

\paragraph{Configuration}
The reward weights are $(\lambda_{\text{tok}},\lambda_{\text{recall}},\lambda_{\text{save}})=(0.3,0.2,0.3)$ and $(\lambda_N,\lambda_T,\alpha)=(0.1,0.3,0.5)$, with no event discount ($\gamma_{\text{seq}}=1$). Retention ratios are requested on $[0.05,1.0]$ with Beta concentration $\kappa_0=8$, so a sampled ratio is $a_i=0.05+0.95\tilde a_i$. Policy and outcome model are trained with Adam at $3\times10^{-4}$ and $10^{-3}$ respectively, with the policy gradient norm clipped at $1$. Appendix~\ref{app:hyperparams} lists the full configuration.

\paragraph{Metrics}
Method comparisons use the paired Wilcoxon signed-rank test. For attribution experiments, the key metric is iso-success token ratio: we compare token ratios only within the success stratum so that savings are never confounded with task failure. Interquartile mean with stratified bootstrap 95\% confidence intervals are also used. The recall column $p_\text{recall}$ in the result tables is the event-local re-invocation rate $r_t$ of Eq.~\ref{eq:precall} taken without the $\lambda_{\text{recall}}$ factor, averaged first over the events of each evaluation record and then over the set $\mathcal{E}$ of valid query--seed records:
\[
p_\text{recall}=\frac{1}{|\mathcal{E}|}\sum_{e\in\mathcal{E}}
  \frac{1}{T_c^{(e)}}\sum_{t=1}^{T_c^{(e)}}r_t^{(e)}.
\]
Statistical details appear in Appendix~\ref{app:stats}.

\begin{table*}[t]
\centering
{\small
\begin{tabular*}{\textwidth}{@{\extracolsep{\fill}}llcccccc@{}}
\toprule
\textbf{Compressor} & \textbf{Arm} & \textbf{Success$\uparrow$} & \textbf{Iso-succ.\ tok.$\downarrow$} & \textbf{95\% CI} & \textbf{$\Delta_\text{tok}$} & \textbf{$p_\text{recall}$$\downarrow$} & \textbf{$\bar{T}_\text{abs}$ (K)}  \\
\midrule
\multirow{2}{*}{Truncation}
  & RL-only  & 74.8\% & 0.728 & {[.703,\,.753]} & --- & 0.022 & 72.8  \\
  & TRACER           & 77.0\% & 0.694 & {[.685,\,.704]} & $-$4.7\% & 0.021 & 69.4  \\
\midrule
\multirow{2}{*}{Summarization}
  & RL-only  & 76.2\% & 0.560 & {[.551,\,.569]} & --- & 0.018 & 47.5  \\
  & TRACER           & \textbf{81.0\%} & \textbf{0.541} & {[.532,\,.550]} & $-$3.4\% & \textbf{0.015} & \textbf{45.9} \\
\midrule
\multirow{2}{*}{Self-Info}
  & RL-only  & 71.5\% & 0.736 & {[.722,\,.750]} & --- & 0.028 & 71.2 \\
  & TRACER           & \textbf{75.0\%} & \textbf{0.706} & {[.693,\,.719]} & $-$4.1\% & \textbf{0.025} & \textbf{68.3} \\
\bottomrule
\end{tabular*}
}
\caption{Ablation of TRACER against the RL-only baseline. Each arm uses 5 seeds}
\label{tab:ablation}
\end{table*}

\begin{figure*}[t]
    \centering
    \includegraphics[width=\textwidth]{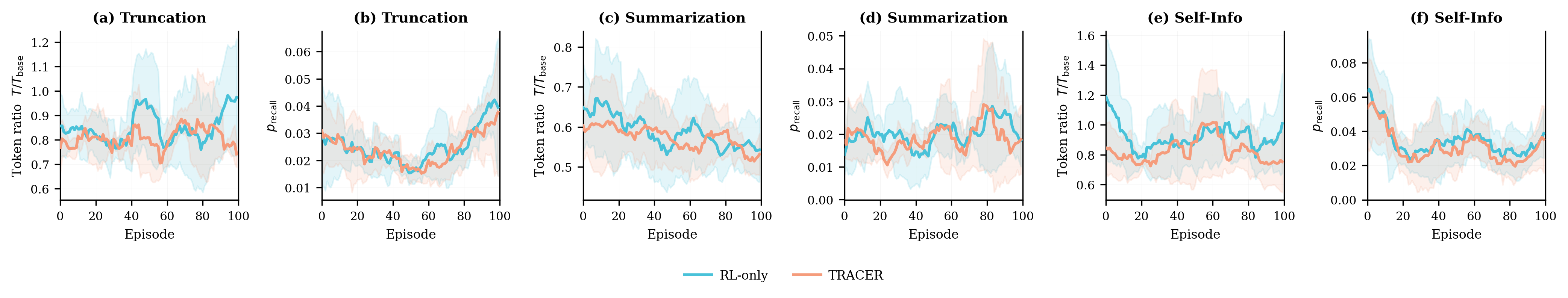}
    \caption{Attribution ablation, RL-only (blue) vs.\ TRACER (orange). Columns are per-compressor; each shows token ratio and re-invocation rate $p_{\mathrm{recall}}$. 5 seeds, 15-episode rolling window; bands are $\pm1$ std.}
    \label{fig:training}
\end{figure*}

\subsection{Main Comparison: TRACER vs.\ Fixed Strategies}
\label{sec:main-results}

\textbf{Does learning per-tool retention ratios improve upon fixed compression strategies?}

Table~\ref{tab:main} shows that fixed baselines largely fail to reduce total tokens. Under truncation and Self-Info, they increase consumption by 3--24\% relative to keep-all, with re-invocation rates at $p_\text{recall}\ge 0.07$. Neither temporal proximity nor output size captures downstream importance. These results directly evidence the compression--consequence gap: indiscriminate compression triggers recovery behavior that offsets or exceeds the initial savings.

TRACER saves 29--46\% of tokens while maintaining task success across all compressors. Summarization benefits most at 46\% because it exploits fine-grained retention signals semantically; Self-Info saves 29\%, truncation 31\%, and the AGORA-style backend 34\%. Re-invocation rates remain at or below 0.025 across all backends, confirming that the learned policy does not purchase savings through costly recovery.

The tool-type-conditional baseline captures part of the inter-tool variance, achieving 11--28\% savings. TRACER further reduces tokens by 15--18\% beyond this static allocation. The residual gap arises because the same tool type requires different retention across queries: an SQL result feeding a downstream multi-step join needs higher retention than one answering a single lookup directly. Query-conditioned adaptation drives the remaining gains. Savings are consistent across difficulty tiers at 32--48\%, with a per-difficulty breakdown in Appendix~\ref{app:difficulty}. Figure~\ref{fig:rl_vs_fixed} visualizes the training dynamics.

\subsection{Ablation: Effect of Attribution}
\label{sec:attribution-results}

\textbf{Does the outcome model improve TRACER's compression decisions?}

\noindent We compare TRACER against an ablated RL-only variant that uses REINFORCE with uniform credit assignment. Both arms share the same policy architecture, reward, and training schedule. The only difference is whether the outcome model provides per-tool directional signals.

Table~\ref{tab:ablation} shows that attribution improves task success across all three compressors: by 4.8~pp under summarization,
  3.5~pp under Self-Info, and 2.2~pp under truncation. The pooled Wilcoxon test confirms significance at $p{=}0.008$. The outcome model
  steers retention toward high-consequence outputs, helping preserve critical information while reducing unnecessary context. This
  reallocation also reduces the iso-success token ratio across all three backends: by 3.4\% under summarization, 4.1\% under Self-Info,
  and 4.7\% under truncation. In addition, the re-invocation probability decreases by 13\% in aggregate. Figure~\ref{fig:training}
  shows that TRACER converges to lower token ratios and re-invocation rates than RL-only.

  Under truncation, attribution increases success by 2.2~pp while reducing the iso-success token ratio from 0.728 to 0.694,
  corresponding to a 4.7\% relative reduction. Although positional truncation offers less flexibility than semantic compression,
  consequence-aware budget allocation can still protect high-consequence outputs while compressing less consequential content more
  aggressively. Attribution therefore provides both success protection and additional token savings on this backend.


\subsection{Validation with Interventional Rollouts}
  \label{sec:counterfactual-validation}

  The ablation shows that attribution improves the policy, but not whether the outcome model correctly predicts the consequence of changing an individual tool's retention ratio. We test this directly against actual single-tool interventions.

  From held-out trajectories we sample compression states $s$ together
  with the frozen policy's action $a=(r_1,\ldots,r_K)$. For each active tool $i$ we replay the agent from the saved snapshot twice: a factual rollout under $a$, and an intervention rollout that sets $r_i{:=}1$ while holding all other ratios fixed. Both use the same frozen deterministic policy after the event and share $M{=}3$ paired seeds to cancel sampling noise. The measured consequence and its model estimate are
  \begin{align}
  \Delta C_i^{\mathrm{roll}}
  &=\tfrac{1}{M}\textstyle\sum_{m=1}^{M}
  \big[C^{(m)}(s,a)-C^{(m)}(s,a;r_i{:=}1)\big],
  \label{eq:true-counterfactual}\\
  \Delta \hat C_i
  &=\hat C(s,a)-\hat C(s,a;r_i{:=}1),
  \label{eq:pred-counterfactual}
  \end{align}
  where a positive value means that fully retaining tool $i$ lowers expected downstream cost relative to the policy-selected ratio.

\begin{table}[h]
  \centering
  {\small
  \setlength{\tabcolsep}{4pt}
  \begin{tabular}{@{}lcccc@{}}
  \toprule
  \textbf{Metric} & \textbf{Trunc.} & \textbf{Summ.} & \textbf{Self-Info} & \textbf{Aggr.} \\
  \midrule
  Spearman $\rho\uparrow$ & $0.64_{\pm.05}$ & $\mathbf{0.73}_{\pm.04}$ & $0.61_{\pm.06}$ & $0.66_{\pm.03}$ \\
  NMAE$\downarrow$        & $\mathbf{0.11}_{\pm.02}$ & $0.12_{\pm.01}$ & $0.17_{\pm.02}$ & $0.13_{\pm.01}$ \\
  Top-3 overlap$\uparrow$ & $0.80_{\pm.05}$ & $\mathbf{0.86}_{\pm.04}$ & $0.81_{\pm.05}$ & $0.82_{\pm.03}$ \\
  Sign Acc.$\uparrow$     & $0.83_{\pm.04}$ & $\mathbf{0.88}_{\pm.03}$ & $0.80_{\pm.04}$ & $0.84_{\pm.02}$ \\
  \bottomrule
  \end{tabular}
  }
  \caption{Validation of the outcome-model counterfactuals against single-tool
  intervention rollouts on held-out states. Cells are
  mean$_{\pm\text{95\% CI}}$.}
  \label{tab:counterfactual-validation}
  \end{table}

  Table~\ref{tab:counterfactual-validation} validates the counterfactual estimates: predicted vs.\ rollout differences yield Spearman
  $\rho=0.66$, the attributor recovers $82\%$ of the true top-3 highest-consequence tools, and $84\%$ of predicted credit directions
  match rollout, confirming the signal provides informative per-tool gradients.

\subsection{Behavioral Analysis}
\label{sec:behavior}

\begin{figure}[h]
    \centering
    \includegraphics[width=\columnwidth]{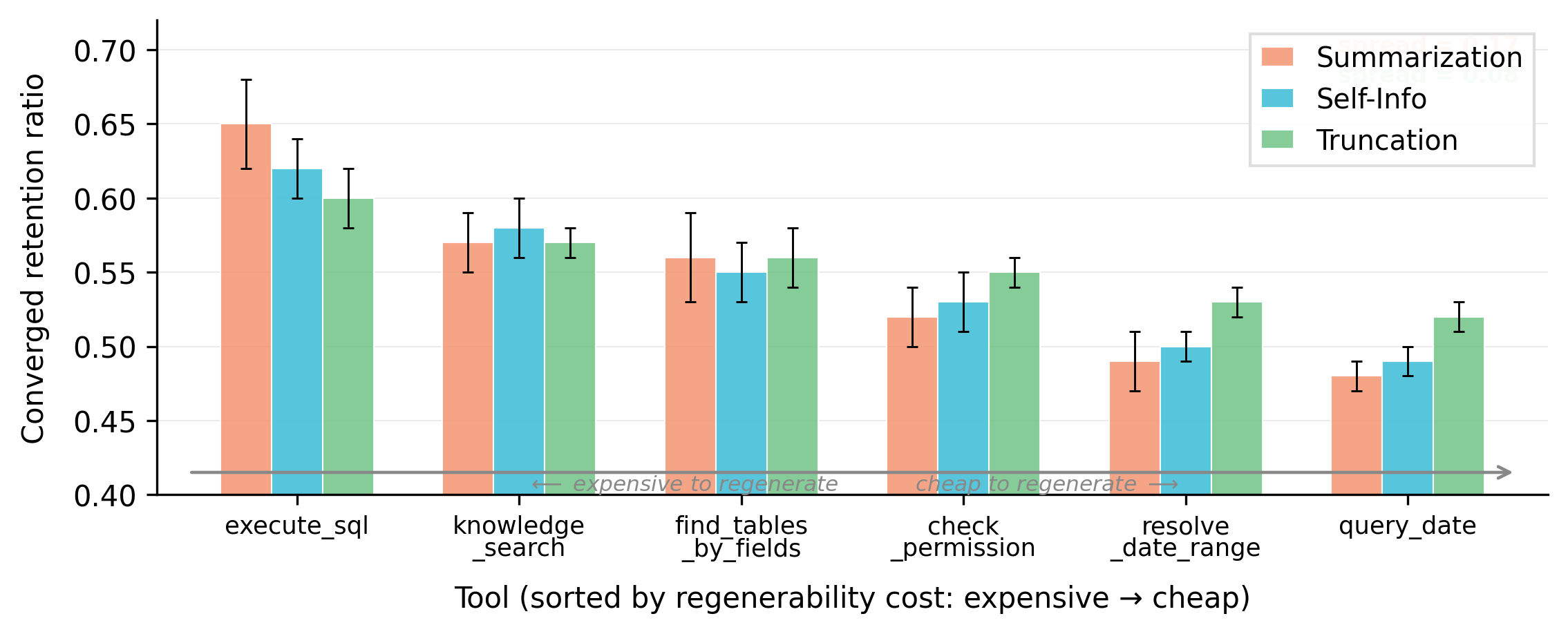}
    \caption{Converged per-tool retention ratios. Tools are sorted by regenerability cost.}
    \label{fig:tool_ratios}
\end{figure}

\paragraph{Per-tool learned ratios}
Figure~\ref{fig:tool_ratios} shows the converged ratios across tools. Under summarization, \texttt{execute\_sql} converges to the highest retention at 0.65, while \texttt{resolve\_date\_range} settles at the lowest at 0.48. This ordering aligns with regenerability cost: SQL results carry irreplaceable numeric data, whereas date lookups return static content that is cheap to regenerate. Under truncation, the spread narrows to 0.52--0.60 because positional truncation cannot exploit fine-grained ratio differences.

\paragraph{Cross-backbone transfer}
To test whether the learned retention structure depends on the training backbone, we freeze the policy trained on Claude Sonnet 4.6 and evaluate it zero-shot on Qwen-3.7-Max~\citep{qwen3} and GPT-5.5~\citep{openai2025gpt55}. Table~\ref{tab:backbone} reports results:

\begin{table}[h]
\centering
{\small
\setlength{\tabcolsep}{3.5pt}
\begin{tabular}{@{}llccc@{}}
\toprule
\textbf{Comp.} & \textbf{Backbone} & \textbf{Succ.$\uparrow$} & \textbf{Tok.\ r.$\downarrow$} & \textbf{$p_\text{rec}$$\downarrow$} \\
\midrule
\multirow{2}{*}{Truncation}
  & Qwen-3.7-Max  & 0.72 & 0.78  & 0.035 \\
  & GPT-5.5       & 0.75 & 0.74  & 0.026 \\
\midrule
\multirow{2}{*}{Summarization}
  & Qwen-3.7-Max  & 0.75 & 0.65  & 0.028 \\
  & GPT-5.5       & 0.78 & 0.60   & 0.020 \\
\midrule
\multirow{2}{*}{Self-Info}
  & Qwen-3.7-Max  & 0.69 & 0.82  & 0.039 \\
  & GPT-5.5       & 0.72 & 0.77  & 0.029 \\
\bottomrule
\end{tabular}
}
\caption{Cross-backbone transfer: frozen policy (trained on Claude Sonnet 4.6) evaluated zero-shot on alternative backbones.}
\label{tab:backbone}
\end{table}

Across both unseen backbones, TRACER retains 18--40\% savings. GPT-5.5 preserves 78--87\% of the training-backbone gains while Qwen-3.7-Max retains 62--76\%, consistent with GPT-5.5's stronger tool-calling compliance. In all reported cases, the frozen TRACER policy achieves positive token
  savings on the two alternative backbones. These results provide initial evidence that the learned retention policy can transfer across model families without retraining.

\paragraph{Deployment cost analysis.}
The one-time training investment (Appendix~\ref{app:hyperparams}) is modest relative to the recurring savings: at production traffic of 1000+ sessions per day, the per-session token reduction recoups the full training cost within 3--6 weeks, after which the learned policy delivers compounding savings indefinitely (Appendix~\ref{app:cost}).

\subsection{Transfer to an AGORA-Inspired Compressor}
\label{sec:agora}

\textbf{Does a policy learned on one compressor transfer to a structurally different one?}

\noindent The three backends above share clause- or position-level granularity. The AGORA-style~\citep{agora2025} step-level compressor is structurally different: it chains clause segmentation, self-information scoring, an always-keep floor for numeric content, and greedy budget fill. To this end, we conduct experiments on Agora-style compressor to demonstrate the generalizability of TRACER.

\begin{table}[h]
\centering
{\small
\begin{tabular}{lccc}
\toprule
\textbf{Strategy} & \textbf{Success$\uparrow$} & \textbf{Tok.\ ratio$\downarrow$} & \textbf{Save} \\
\midrule
Uniform-0.25              & 0.65 & 1.208 & $-$21\% \\
TRACER (zero-shot)        & 0.62 & 0.690 & 31\% \\
\textbf{TRACER (on AGORA)} & \textbf{0.65} & \textbf{0.656} & \textbf{34\%} \\
\bottomrule
\end{tabular}
}
\caption{Transfer to the AGORA-style compressor.}
\label{tab:agora}
\end{table}

From table~\ref{tab:agora}  we can observe that uniform-0.25 increases total tokens by 21\% relative to keep-all: its aggressive budget forces the always-keep floor to dominate, leaving insufficient budget for non-floor clauses and triggering recovery calls.

Also, table~\ref{tab:agora} reports two TRACER variants. The first, trained only on the truncation backend and applied zero-shot, achieves 31\% savings at success 0.62. The second, trained directly on the AGORA backend, reaches 34\% savings at matched success 0.65 by learning per-tool budgets that cooperate with the floor mechanism. The per-tool importance structure TRACER learns transfers across backends without fine-tuning, and target-specific training recovers the remaining gap.

\subsection{Generalization on LOCA-bench}
\label{sec:loca}

To further test the generalization, we evaluate TRACER on LOCA-bench~\citep{locabench2026}, a public benchmark with 15 task environments and deterministic workspace-comparison evaluation. The tool set and task distribution are entirely disjoint from our production training environment. We use the three longest context-window settings (96K, 128K, 256K) where compression pressure is strongest. We train a fresh TRACER policy via 5 rounds of iterative RL over 10 task environments and evaluate on 5 held-out environments. Table~\ref{tab:loca} reports success rates aggregated across the three context windows.

\begin{table}[h]
\centering
{\small
\begin{tabular}{llcc}
\toprule
\textbf{Compressor} & \textbf{Strategy} & \textbf{Success$\uparrow$} & \textbf{Tok.\ ratio$\downarrow$}  \\
\midrule
\multicolumn{2}{l}{Keep-all} & 0.37 & 1.000  \\
\midrule
\multirow{3}{*}{Truncation}
  & Recency        & 0.39 & 0.978  \\
  & Token-prop     & 0.40 & 0.961  \\
  & \textbf{TRACER} & \textbf{0.44} & \textbf{0.751}  \\
\midrule
\multirow{3}{*}{Summarization}
  & Recency        & 0.41 & 0.847  \\
  & Token-prop     & 0.23 & 0.938  \\
  & \textbf{TRACER} & \textbf{0.48} & \textbf{0.806} \\
\midrule
\multirow{3}{*}{Self-Info}
  & Recency        & 0.35 & 0.891  \\
  & Token-prop     & 0.24 & 0.922  \\
  & \textbf{TRACER} & \textbf{0.38} & \textbf{0.815} \\
\bottomrule
\end{tabular}
}
\caption{Generalization on LOCA-bench (5 held-out tasks $\times$ 5 seeds $\times$ 3 context windows).}
\label{tab:loca}
\end{table}

From table~\ref{tab:loca} we can know that TRACER outperforms keep-all across all compressor backends. TRACER achieves higher aggregate success than keep-all in the reported long-context settings. Also, TRACER outperforms heuristic baselines while achieving token savings in general. Since uniform-0.5 suffers from truncation that is too aggressive and inflexible, it almost never succeeds on LOCA-bench; we therefore omit further discussion of it here. To be more specific, TRACER's gains concentrate on data-analysis tasks for it preserves data-bearing outputs and compresses auxiliary ones. It also works well on procedural tasks with strong sequential dependencies because its policy identifies the causally critical nodes in the operation chain and allocates context budget to them, allowing the agent to retain a coherent execution spine under severe pressure and complete workflows that other strategies cannot.

\section{Conclusion}
\label{sec:conclusion}

We presented TRACER, a reinforcement-learning framework that treats context compression as a sequential per-tool decision problem. Our method saves tokens across four compressor backends while preserving task success. A two-channel credit scheme attributes consequences to individual tools through single-tool counterfactuals, lifting success on semantic compressors. The learned policy generalizes to held-out queries, transfers zero-shot across agent backbones and compressor architectures, and achieves 18--25\% savings on LOCA-bench. 

\bibliography{paper_refs}

\clearpage
\section*{Technical Appendix}
\appendix
\setcounter{table}{0}
\renewcommand{\thetable}{A\arabic{table}}

\section{Score Function of the Beta Policy}
\label{app:beta}
The reference implementation uses $a_{\min}=0.05$, $a_{\max}=1.0$, and $\kappa_0=8$. Thus $a_i=0.05+0.95\tilde a_i$, where $\tilde a_i\in(0,1)$ has mean $\tilde m_i=\sigma(\text{net}_i)$, with $\alpha_i=8\tilde m_i$ and $\beta_i=8(1-\tilde m_i)$. The log-density is
\begin{equation*}
\begin{aligned}
\log\pi(a_i\mid s)={}&(\alpha_i{-}1)\log\tilde a_i+(\beta_i{-}1)\log(1{-}\tilde a_i)\\
&-\log B(\alpha_i,\beta_i)-\log 0.95.
\end{aligned}
\end{equation*}
Differentiating $\log B$ gives digamma terms $\partial_{\alpha}\log B=\dgm(\alpha)-\dgm(\alpha+\beta)$ and $\partial_{\beta}\log B=\dgm(\beta)-\dgm(\alpha+\beta)$. Since $\partial\alpha_i/\partial\tilde m_i=\kappa_0$ and $\partial\beta_i/\partial\tilde m_i=-\kappa_0$, the two $\dgm(\alpha_i+\beta_i)$ terms cancel:
\begin{equation}
\frac{\partial\log\pi}{\partial\tilde m_i}=\kappa_0\Big[\logit(\tilde a_i)-\big(\dgm(\alpha_i)-\dgm(\beta_i)\big)\Big]=:s_i.
\label{eq:score}
\end{equation}
Using the Beta log-moment identity $\Ex[\log X]=\dgm(\alpha)-\dgm(\alpha+\beta)$ (differentiate the normalizer under the integral), $\Ex[\logit\tilde a_i]=\dgm(\alpha_i)-\dgm(\beta_i)$, so $\Ex[s_i]=0$. Zero mean makes any constant or action-independent baseline unbiased, and gives $g_i=\Ex[\text{credit}_i\,s_i]=\Cov(\text{credit}_i,s_i)$, the entry point for Appendix~\ref{app:sign}. At evaluation, the deterministic Beta mean is mapped back to the retention interval, so the applied ratio is $a_i=0.05+0.95\tilde m_i$.

\section{Full Training Algorithm}
\label{app:algorithm}

Algorithms~\ref{alg:tracer} and~\ref{alg:tracer-update} give the event-level training path implemented in the supplement. Phase~0 keep-all rollouts establish $T_{\text{base}}$ but are not inserted into the outcome-model replay buffer. The coevolve arm initializes its outcome model randomly and does not accept a pretrained predictor. Eligible policy-rollout events accumulate during the $\beta=0$ warmup; the first fit occurs immediately at the boundary after the 20th completed episode, and later fits occur every five completed episodes. Consequently, $r_i{:=}1$ in Eq.~(7) is a model counterfactual, not a separately observed Phase~0 label.

\begin{algorithm*}[t]
\caption{TRACER: Rollout and Event Recording}
\label{alg:tracer}
\begin{algorithmic}[1]
\REQUIRE Queries $\mathcal{Q}$, compressor $\mathcal{C}$, agent $\mathcal{A}$, per-run trigger $\tau$ on backend-estimated full $|c_t|$
\STATE Initialize policy $\pi_\theta$ ($174\to128\to64\to N$ MLP), random outcome model $f_\phi$, and replay buffer $\mathcal{B}$ (capacity 500)
\STATE Initialize per-query EMA baselines $b_q$ and reward histories $\mathcal{H}_q$
\STATE \textbf{Phase 0: Baseline Collection}
\FOR{each $q \in \mathcal{Q}$}
    \STATE Run $\mathcal{A}$ five times with $r_i{=}1$; set $T_{\text{base}}(q)$ to mean total tokens and retain reference timelines
\ENDFOR
\STATE \textbf{Phase 1: Online Training}
\FOR{episode $e = 1, \dots, E$}
    \STATE Sample query $q \sim \text{Uniform}(\mathcal{Q})$
    \STATE $\beta_e\gets0$ for $e\le20$; otherwise $\beta_e\gets\min\{1,(e-20)/10\}$
    \STATE Initialize event records $\mathcal{D}\gets\emptyset$
    \STATE Run $\mathcal{A}$ on $q$ with the compression policy
    \FOR{each compression event $t = 1, \dots, T_c$}
        \STATE Build ordered output-instance slots, state $s_t$, and the active mask
        \STATE $\tilde m_i \gets \sigma(\text{MLP}_\theta(s_t)_i)$ for each tool slot $i$
        \STATE Sample $\tilde a_{t,i}\sim\text{Beta}(8\tilde m_i,8(1-\tilde m_i))$; request $r^{\mathrm{req}}_{t,i}\gets0.05+0.95\tilde a_{t,i}$
        \STATE Apply $\mathcal{C}$ with requested ratios $\{r^{\mathrm{req}}_{t,i}\}$
        \STATE Observe $N_{\text{post},t}$ and matched keep-all count $N_{\text{pre},t}$ in the same event window
        \STATE Compute event-local $r_t$ with the Eq.~(2) zero-reference rule; set $P_{\text{recall},t}\gets\lambda_{\text{recall}}r_t$
        \STATE Append $(s_t,a_t{:=}r^{\mathrm{req}}_t,r_t,P_{\text{recall},t},\text{active}_t,\text{event targets})$ to $\mathcal{D}$
    \ENDFOR
    \IF{any event-local recall measurement is missing}
        \STATE Skip the update rather than treating missing recall as zero
    \ENDIF
    \STATE Observe success $\in\{0,1\}$ and total session tokens $T$
    \STATE $P_{\text{recall}}\gets\sum_{t=1}^{T_c}P_{\text{recall},t}$ \hfill (Eq.~3)
    \STATE $R_{\text{base},T_c}\gets\text{success}-\lambda_{\text{tok}}\clamp(T/T_{\text{base}},0.2,2)$; $R_{\text{base},k<T_c}\gets0$ \hfill (Eq.~1)
    \STATE $\sigma_q\gets1$ if $|\mathcal{H}_q|<3$, else $\max(\operatorname{std}(\mathcal{H}_q[-20:]),0.01)$
    \STATE $A\gets(R_{\text{base},T_c}-b_q)/\sigma_q$; update $b_q$ with decay 0.9 and append $R_{\text{base},T_c}$ to $\mathcal{H}_q$
    \STATE Pass $(e,\beta_e,\mathcal{D},A,T_c,T_{\text{base}})$ to Algorithm~\ref{alg:tracer-update}
\ENDFOR
\RETURN Frozen policy $\pi_{\theta^*}$; use its deterministic mean for evaluation
\end{algorithmic}
\end{algorithm*}

\begin{algorithm*}[t]
\caption{TRACER: Policy and Outcome-Model Updates}
\label{alg:tracer-update}
\begin{algorithmic}[1]
\REQUIRE Episode $e$, attribution weight $\beta_e$, records $\mathcal{D}$, advantage $A$, event count $T_c$, baseline $T_{\text{base}}$
    \FOR{each event record $t\in\mathcal{D}$}
        \STATE $A_t\gets\gamma_{\text{seq}}^{T_c-t}A=A$ with $\gamma_{\text{seq}}=1$
        \STATE $\text{credit}^{\text{floor}}_{t,i}\gets-P_{\text{recall},t}/|\text{active}_t|$ for $i\in\text{active}_t$ \hfill (Eq.~6)
        \STATE $\text{credit}^{\text{attr}}_{t,i}\gets0$
        \IF{$\beta_e > 0$}
            \STATE $\hat C(s_t,a_t)\gets\lambda_N\sum_j\hat n_j+2\lambda_T\tanh\!\left(\frac{\exp\hat T_{\log}}{2T_{\text{base}}}\right)$
            \STATE $\Delta\hat C_{t,i}\gets\max\{0,\hat C(s_t,a_t)-\hat C(s_t,a_t;r^{\mathrm{req}}_{t,i}{:=}1)\}$
            \STATE $\text{credit}^{\text{attr}}_{t,i}\gets\lambda_{\text{save}}(1-r^{\mathrm{req}}_{t,i})\frac{\text{tok}_{t,i}}{T_{\text{base}}}-\Delta\hat C_{t,i}$ \hfill (Eq.~7)
        \ENDIF
        \STATE $\text{coeff}_{t,i}\gets\alpha A_t+\text{credit}^{\text{floor}}_{t,i}+\beta_e\,\text{credit}^{\text{attr}}_{t,i}$ \hfill (Eq.~8)
    \ENDFOR
    \STATE \textbf{Policy Update (shaped surrogate):}
    \STATE $\mathcal{L} \gets -T_c^{-1}\sum_t\sum_{i \in \text{active}_t} \text{coeff}_{t,i}\log\pi_\theta(a_{t,i}\mid s_t)$
    \STATE Adam update of $\theta$ (LR $3\times10^{-4}$; gradient norm clipped at 1)
    \STATE Add eligible event records with propensity and outcomes to $\mathcal{B}$; mark episode $e$ completed
    \IF{$e=20$ or ($e>20$ and $(e-20)\bmod5=0$)}
        \STATE Update $f_\phi$ for 10 steps from batches of 16 using the multi-task loss in Appendix~\ref{app:hyperparams}
        \STATE At $e=20$, fail the coevolve run if eligible labels are insufficient or the first fit fails
    \ENDIF
    \STATE Every 10 episodes, retain the checkpoint with the best rolling-10 mean reward
\end{algorithmic}
\end{algorithm*}

\section{Gradient-Direction Intuition for Attribution}
\label{app:sign}

This section provides implementation intuition rather than a new theoretical claim. Equation~(7) combines two terms with opposing effects. The immediate-saving term
$\lambda_{\text{save}}(1-r_i^{\mathrm{req}})\text{tok}_i/T_{\text{base}}$ uses the ratio sampled and requested by the policy; it grows as requested retention decreases and therefore rewards compression. The counterfactual downstream-cost term is $[\hat C(s,a)-\hat C(s,a;r_i^{\mathrm{req}}{:=}1)]_+$, so only predicted harm relative to fully retaining tool $i$ is subtracted. Predicted negative cost differences do not create an extra reward. The net credit favors compression for low-consequence outputs and protects outputs with predicted downstream cost.

The model is not assumed to provide a formal causal guarantee. Its usefulness is evaluated empirically by the attribution ablation in Table~2 and the single-tool interventional rollouts in Table~3 of the main paper.

\section{Sequential Semi-MDP Formulation}
\label{app:semimdp}
With events $t=1,\dots,T_c$, the main paper defines the base reward only at the terminal event. No discount factor is specified or needed to restate that objective. The undiscounted base return and normalized advantage used by the shaped surrogate are
\begin{align*}
G_t^{\text{base}}&=\sum_{k=t}^{T_c}R_{\text{base},k}=R_{\text{base},T_c},\\
A_t&=\frac{G_t^{\text{base}}-b_q}{\sigma_q},\\
g_\theta^{\text{shaped}}&=\Ex\Big[\sum_t\sum_{i\in\text{active}_t}\text{coeff}_{i,t}\nabla_\theta\log\pi_\theta(a_{t,i}\mid s_t)\Big].
\end{align*}
Because $\alpha$ rescales $A_t$ and $\text{credit}^{\text{attr}}_{i,t}$ adds model-based shaping, $g_\theta^{\text{shaped}}$ is the expected update direction of Eqs.~(5) and (8), not an unbiased estimator of $\nabla_\theta J$ for the task-level objective in Eq.~(3). $P_{\text{recall},t}$ stays event-local and is consumed by the credit at event $t$; it is not replaced by a session-level count. Reference windows are obtained by replaying the same compression-trigger boundaries over the five keep-all timelines and matching windows by event order; $N_{\text{pre},t}$ is their mean call count. The active mask sums log-probabilities over all tools present at the event, so every event is logged separately. The reference implementation estimates $\sigma_q$ from the most recent 20 rewards after at least three observations, with a floor of $0.01$, as specified in Algorithm~\ref{alg:tracer} and Table~\ref{tab:hyperparams}.

\section{Provenance-Based Attribution}
\label{app:prov}
This appendix specifies the provenance upgrade referenced in the main paper. It replaces the proximal time-window proxy with causal source matching: each removed span $\sigma$ at event $t$ is tagged with $\langle t, i(\sigma), \phi(\sigma)\rangle$, where $\phi$ is a content signature, and a later tool call $\rho$ is matched to an earlier removed span:
\begin{equation*}
\begin{aligned}
\mathrm{match}(\rho)=\argmax_{\substack{\sigma:\,\mathrm{time}(\sigma)<\mathrm{time}(\rho)}}&\mathrm{sim}(\rho,\phi(\sigma))\\[-2pt]
&\text{if}\ \max\mathrm{sim}\ge\tau_{\mathrm{prov}},\ \text{else}\ \varnothing.
\end{aligned}
\end{equation*}
The resulting count $n^{\text{prov}}_{t,i}$ sums matched costs and defines a provenance-based penalty and targeted floor:
\begin{equation*}
\begin{aligned}
\text{credit}^{\text{floor-prov}}_i&=-P^{\text{prov}}_{\text{recall},t}\cdot\frac{n^{\text{prov}}_{t,i}}{\sum_j n^{\text{prov}}_{t,j}},\\
\textstyle\sum_i\text{credit}^{\text{floor-prov}}_i&=-P^{\text{prov}}_{\text{recall},t}.
\end{aligned}
\end{equation*}
When $\sum_j n^{\text{prov}}_{t,j}>0$, the floor conserves the penalty by construction and removes the time-window mixing identified in the main paper. Match precision remains an empirical property of the signature and threshold, so a deployment should calibrate hit and mismatch rates on labeled calls. The reported experiments and Algorithm~\ref{alg:tracer} retain the measured window-based signal in Eq.~(2); provenance is the stated upgrade path rather than a replacement for those reported numbers.

\section{Design Considerations and Solutions}
\label{app:defects}
\begin{itemize}
    \item \textbf{Bounded action.} The Beta policy in Eq.~(4) of the main text matches the bounded action support without post-sampling truncation.
    \item \textbf{Smooth downstream cost.} The reported $\hat C$ uses $2\tanh(x/2)$ for its token term. No result for a hard-clamped alternative is reported.
    \item \textbf{Optional inverse-propensity weighting.} A stabilized self-normalized inverse-propensity weight could be investigated when action--state dependence is a concern. It is not part of the objective, algorithm, or reported experiments, and this appendix makes no identifiability claim for it.
    \item \textbf{Optional provenance matching.} The measured proximal-window target can include natural call-count variation. Appendix~\ref{app:prov} describes a possible upgrade, not a component of the reported method.
\end{itemize}

\section{Feature Definitions}
\label{app:features}

The released policy state is
\[
s_t=[q_1,\ldots,q_7;\ o_{1,1},\ldots,o_{16,10};\ c_1,\ldots,c_7]\in\Real^{174}.
\]
The 16 positions are fixed-width tensor slots for ordered tool-output instances present at the current compression event, not one slot per canonical tool type. Repeated calls to the same tool retain distinct output identifiers, slots, ratios, and gradients. Unused positions are zero-padded and masked. The reference implementation rejects events containing more than 16 eligible outputs rather than silently merging or dropping them.

\begin{table*}[t]
\centering
\small
\begin{tabular}{@{}llp{0.70\textwidth}@{}}
\toprule
\textbf{Block} & \textbf{Dim.} & \textbf{Meaning} \\
\midrule
Query & $q_1$ & $0.3\log(1+|q|/2)$, a token-length proxy. \\
 & $q_2$ & Numeric-constraint indicator. \\
 & $q_3$ & Multi-hop/comparison indicator. \\
 & $q_4$--$q_7$ & Mutually exclusive retrieval, SQL/aggregation, report, and other intents. \\
\midrule
Output $i$ & $o_{i,1}$ & Presence flag. \\
 & $o_{i,2}$ & $0.1\log(1+\text{tok}_i)$, output-size feature. \\
 & $o_{i,3}$ & Output-token share within the current event. \\
 & $o_{i,4}$ & $0.1\log(1+\text{same-type tokens})$ within the event. \\
 & $o_{i,5}$ & $\log(1+\text{same-type ordinal})$ for repeated calls. \\
 & $o_{i,6}$ & Query relevance; fallback static criticality prior high/medium/low $=1.0/0.5/0.2$. \\
 & $o_{i,7}$ & Prefix-available downstream-reuse annotation, else zero. \\
 & $o_{i,8}$ & Prefix-available redundancy annotation, else zero. \\
 & $o_{i,9}$ & Normalized event-slot recency, from oldest 0 to newest 1. \\
 & $o_{i,10}$ & Prefix rework annotation when available, else zero. \\
\midrule
Context & $c_1$--$c_5$ & System, instruction/tool-definition, dialogue-history, tool-output, and scratchpad fractions. \\
 & $c_6$ & Context pressure relative to compression threshold $\tau$. \\
 & $c_7$ & Normalized compression-event index. \\
\bottomrule
\end{tabular}
\caption{Reference implementation feature specification. Runtime annotations use only values available in the trajectory prefix; unavailable annotations default to zero or the stated prior.}
\label{tab:features}
\end{table*}

The outcome model reuses the ten dimensions for each output and appends the applied retention ratio as an eleventh dimension. A separate seven-dimensional query token is projected into the same 64-dimensional embedding space. It predicts $\hat n_i$, $\hat T_{\log}$, and the auxiliary success probability $\hat p$; only $\hat n_i$ and $\hat T_{\log}$ enter $\hat C$.

\section{Reported Baselines and Experimental Configuration}
\label{app:hyperparams}

\subsection{Baseline Strategy Formulas}
\label{app:baselines}

We restate the baseline definitions from Section~4.1 of the main text. Exact formulas are shown only when the main text specifies them.

\paragraph{Keep-all.} $r_i = 1$ for all tools. Serves as the success ceiling and token baseline ($T_\text{base}$).

\paragraph{Uniform-0.5.} $r_i = 0.5$ for all tools, regardless of type, size, or position.

\paragraph{Recency.} The ratio decays linearly from the most recent tool output (highest ratio) to the oldest:
\begin{equation*}
r_i = 0.1 + 0.8 \cdot \frac{\text{pos}_i - 1}{N - 1},
\end{equation*}
where $\text{pos}_i \in \{1, \dots, N\}$ is the position of tool $i$ sorted by invocation time (most recent = $N$). This yields $r = 0.9$ for the most recent and $r = 0.1$ for the oldest.
For the single-output case $N=1$, the reference implementation uses the midpoint $r=0.5$.

\paragraph{Token-proportional.} For output-size estimate $\text{size}_i$,
\[
r_i=1-0.7\frac{\text{size}_i}{\max_j\text{size}_j}.
\]
The largest output receives 0.3; smaller outputs approach 1.

\paragraph{Tool-type-conditional.} The released search uses greedy coordinate descent over $\{0.2,0.3,\ldots,0.9\}$, visiting canonical tool types in decreasing mean-output-size order. Each coordinate maximizes success rate minus $0.3$ times token ratio on training queries while holding other coordinates fixed; the default script runs two passes.

\subsection{Reference Implementation Configuration}

Table~\ref{tab:hyperparams} records the checked-in defaults used by the reference implementation. These values document the released training path; they are not a per-table run manifest, and command-line or environment overrides must be recorded when regenerating an experiment.

\begin{table*}[t]
\centering
\small
\begin{tabular}{@{}llp{0.38\textwidth}@{}}
\toprule
\textbf{Group} & \textbf{Setting} & \textbf{Value} \\
\midrule
Reward & $\lambda_{\text{tok}},\lambda_{\text{recall}},\lambda_{\text{save}}$ & $0.3,0.2,0.3$ \\
 & $\lambda_N,\lambda_T,\alpha$ & $0.1,0.3,0.5$ \\
 & $\gamma_{\text{seq}}$ & $1.0$ (no event discount) \\
Policy & Architecture & $174{\to}128{\to}64{\to}16$, ReLU \\
 & Action interval; $\kappa_0$ & $[0.05,1.0]$; $8.0$ \\
 & Optimizer; LR; grad clip & Adam; $3\times10^{-4}$; 1.0 \\
Baseline & EMA decay & $0.9$ \\
 & Scale estimate & Last 20 rewards after 3 samples; floor $0.01$ \\
Outcome & Architecture & $d=64$, 2 layers, 4 heads, FF 128, dropout 0.1 \\
 & Inputs & 16 output tokens $\times$ 11 dims; one 7-dim query token \\
 & Optimizer; LR & Adam; $10^{-3}$ \\
 & Loss weights $(w_n,w_T,w_s)$ & $(1.0,1.0,0.3)$ \\
 & Monotonicity weight & $\lambda_{\text{mono}}=0.5$ \\
 & Batch; replay capacity & 16; 500 event records \\
 & Refresh work & 10 gradient steps per refresh \\
 & EMA anchor & Decay 0.995 after warmup \\
Schedule & Warmup; annealing; refresh & 20 episodes; 10 episodes; at 20 completed episodes, then every 5 \\
Environment & Compression threshold & Caller-supplied per run; backend-estimated full $|c_t|$ \\
 & Keep-all reference & 5 runs per query \\
 & Training length; seeds & 100 episodes; $\{0,1,2,3,4\}$ \\
\bottomrule
\end{tabular}
\caption{Reference implementation defaults.}
\label{tab:hyperparams}
\end{table*}

The outcome loss is
\begin{equation*}
\begin{aligned}
\mathcal{L}_{\phi}
={}&\operatorname{MSE}(\hat n,n)
  +\operatorname{MSE}(\hat T_{\log},\log T)\\
 &+0.3\operatorname{BCE}(\hat p,\text{success})
  +0.5\Omega_{\text{mono}},
\end{aligned}
\end{equation*}
with count and token predictions and targets standardized by running statistics. The auxiliary success head is trained by this loss but is omitted from $\hat C$. The released updater can additionally apply propensity weighting and an EMA anchor; these are implementation stabilizers rather than new terms in the paper objective.

\subsection{Computing Infrastructure and Runtime}
\label{app:infra}

The reference environment used a single Intel Xeon Platinum 8369B server (32 CPU cores, 128\,GB RAM) running Alibaba Cloud Linux~3 with kernel~5.10. The software stack was Python~3.12.4, PyTorch~2.3.1, NumPy~1.26.4, SciPy~1.13.0, Flask~3.0.3, and Transformers~4.42.0. The policy and outcome model together contain 102{,}306 trainable parameters (31{,}696 and 70{,}610, respectively), approximately 102K, and run on CPU; external agent and compressor-model calls dominate wall-clock cost.

The production agent backbone is Claude Sonnet~4.6 with a 200K-token context window. Truncation is local string slicing, Self-Info uses a local Qwen-2.5-0.5B clause scorer, and summarization uses Qwen-3.7-Max through an API; the AGORA-inspired backend combines step/clause segmentation, self-information scoring, and a numeric-content floor. A rollout takes roughly 30--180 seconds depending on query and backend. A 100-episode, five-seed run for one compressor takes about 25--42 hours. The broader experimental campaign was approximately 400 compute-hours and \$2{,}800 in backbone API charges; these figures are operational measurements, not controlled efficiency outcomes.

\subsection{Outcome Model Held-Out Accuracy}
\label{app:outcome-accuracy}

Table~\ref{tab:outcome-accuracy} restates the two held-out metrics reported in Section~3.4 of the main text. 

\begin{table*}[t]
\centering
\small
\begin{tabular}{@{}l c p{5.2cm}@{}}
\toprule
\textbf{Metric} & \textbf{Value} & \textbf{Description} \\
\midrule
Spearman $\rho$ ($\hat n_i$) & 0.72 & Rank correlation between predicted and actual per-tool re-invocation counts \\
Brier score ($\hat p$) & 0.07 & Calibration of success-probability prediction \\
\bottomrule
\end{tabular}
\caption{Outcome-model held-out metrics reported in the main paper.}
\label{tab:outcome-accuracy}
\end{table*}

The Spearman result supports the directional ranking used by the counterfactual credit. The success-probability head $\hat p$ is an auxiliary prediction head evaluated by the Brier score. It is excluded from the downstream-cost definition $\hat C(s,a)=\lambda_N\sum_j\hat n_j+2\lambda_T\tanh\!\left(\exp(\hat T_{\log})/(2T_{\text{base}})\right)$ in the main paper.

\subsection{LOCA-bench Training Configuration}
\label{app:loca-training}

Section~4.7 and Table~6 of the main paper report the LOCA-bench experiment. A fresh TRACER policy is trained for five rounds of iterative RL on 10 of the benchmark's 15 task environments and evaluated on the other 5. Evaluation aggregates the 96K, 128K, and 256K context-window settings over 5 seeds.

\section{Difficulty Tiers and Evaluation Boundary}
\label{app:difficulty}

The full production-query corpus contains 120 queries in three tiers:
\begin{itemize}
    \item 25 simple lookups;
    \item 55 multi-step queries combining schema retrieval with SQL; and
    \item 40 multi-table permission-gated joins.
\end{itemize}

The stratified 80/40 split yields the following held-out composition:
\begin{table*}[t]
\centering
\small
\begin{tabular}{@{}lccc@{}}
\toprule
\textbf{Tier} & \textbf{Simple} & \textbf{Multi-step} & \textbf{Multi-table} \\
\midrule
Held-out queries & 7 & 19 & 14 \\
Reported saving & \multicolumn{3}{c}{32--48\% across the reported tier--compressor cells} \\
\bottomrule
\end{tabular}
\caption{Composition of the held-out production-query split.}
\label{tab:held-out-composition}
\end{table*}
Held-out evaluation uses frozen deterministic policies on these 40 queries. The main paper reports the cross-cell range above.

\section{Statistical Procedures and Reporting Boundary}
\label{app:stats}

Each RL configuration trains for 100 episodes per seed with five seeds; this training schedule is not itself the sample size of a held-out hypothesis test. Test sample sizes are determined after aligning the available method records on their observation keys and applying the metric-specific validity filter. No post-hoc power calculation is reported.

\subsection{Reported Recall Statistic}

The uppercase $P_{\mathrm{recall},t}=\lambda_{\mathrm{recall}}r_t$ in
Eq.~(2) is a training penalty, where the event-local rate is
\[
r_t^{(e)}=\clamp\!\left(
  \frac{N_{\mathrm{post},t}^{(e)}-N_{\mathrm{pre},t}^{(e)}}
       {N_{\mathrm{pre},t}^{(e)}},0,2\right).
\]
For the set $\mathcal{E}$ of valid query--seed evaluation records, the
lowercase table statistic first averages events within each record and then
averages the resulting record-level rates:
\[
p_{\mathrm{recall}}
=\frac{1}{|\mathcal{E}|}\sum_{e\in\mathcal{E}}
  \frac{1}{T_c^{(e)}}\sum_{t=1}^{T_c^{(e)}}r_t^{(e)}.
\]
Zero-reference cases are handled as in Algorithm~\ref{alg:tracer}. Thus
$p_{\mathrm{recall}}$ is an event-local rate, not a probability or a training
penalty: it does not include $\lambda_{\mathrm{recall}}$ and does not sum
across events. Training curves apply the stated rolling-window mean to the
corresponding record-level rates.

\subsection{Confidence Interval Construction}

Method-comparison token summaries use the interquartile mean (IQM) with 10{,}000 stratified bootstrap resamples. Observations are grouped by difficulty tier; within each resample, observations are sampled with replacement inside each tier while preserving the original tier sizes. The 2.5th and 97.5th percentiles of the bootstrap IQM distribution form the 95\% interval. If no stratification variable is supplied, the implementation uses a single stratum rather than treating each observation as its own stratum.

Table~3 uses a different construction for counterfactual validation: each cell center is the metric's native point estimate, and percentile-bootstrap resampling treats each held-out state as a cluster so that paired interventions from the same state remain together. The reported $\pm$ value is the half-width of the resulting 95\% interval. These Table~3 intervals are not IQM intervals.

\subsection{Hypothesis Testing}

Method comparisons use a two-sided paired Wilcoxon signed-rank test. Records are aligned on the common explicit observation key---normally $(\text{query id},\text{seed},\text{episode})$ for training histories and $(\text{query id},\text{seed})$ for one-shot held-out runs---before testing; unmatched records are not positionally paired. The pooled attribution-ablation result reported in the main paper is $p=0.008$. Table~1 has no $p$-value column, so the appendix does not attribute unreported per-baseline $p$-values to it. Cliff's $\delta$ and Holm--Bonferroni utilities are included for additional analyses but are not retroactively attached to the main table.

\paragraph{Success conditioning.}
For the attribution experiments, the main paper defines the key token metric as the iso-success token ratio, computed within the success stratum. That conditioning should not be silently extended to every token ratio in the paper: the main comparison defines token ratio simply as $T/T_{\text{keep-all}}$.

\subsection{Reproducibility of Randomness}

Production-query RL runs use seeds $\{0,1,2,3,4\}$. Each run seeds Python \texttt{random}, NumPy, and PyTorch before query sampling, policy initialization, Beta exploration, and replay-buffer sampling; output names carry the seed suffix. Frozen evaluation uses the deterministic policy mean. The LOCA launcher additionally propagates its selected seed through \texttt{PYTHONHASHSEED}. These controls reproduce the released program's stochastic choices but cannot make remote model or production-backend responses deterministic.

\section{Extended Related Work}
\label{app:extended-related}

Section~2 of the main paper groups related approaches into step/turn-level credit~\citep{wang2026steppo,li2025turnppo,wang2026ecpo}, structural or process-reward decomposition~\citep{cheng2026graphgpo,lu2026hisr,su2025eapo,cm2-2026,yuan2026vpr,wang2026rethinking}, and redundant-tool-call penalties~\citep{toolaware2026,learnwhennot2026}. These approaches optimize the agent's action policy, whereas TRACER freezes that policy and trains the context policy.

TRACER draws on counterfactual credit assignment~\citep{foerster2018counterfactual,wolpert2001optimal,arjona2019rudder,chen2025emai}: it compares the predicted consequence of the selected retention ratio with a single-tool fully retained alternative. This is the distinction established by the main paper; no additional claims about the cited methods' implementations are needed for reproduction.

\section{Cost-Benefit Analysis}
\label{app:cost}

Training adds an up-front cost, whereas token savings recur with use. For a deployment with training cost $C_{\text{train}}$, average keep-all tokens $T_0$, observed token ratio $r$, token price $p$, and daily volume $V$, a simple scenario calculation is
\begin{equation*}
d_{\text{break-even}}=\frac{C_{\text{train}}}{V\,T_0(1-r)p}.
\end{equation*}
The point calculation underlying the deployment discussion uses the conservative full-campaign backbone-API expenditure $C_{\text{train}}=\$2{,}800$, $T_0=85{,}000$ tokens, the Summarization ratio $r=0.541$, an input-token price $p=\$3/10^6$ tokens, and $V=1{,}000$ sessions/day. It gives
\begin{align*}
\Delta T_{\text{session}}&=85{,}000(1-0.541)=39{,}015\ \text{tokens},\\
s_{\text{session}}&=39{,}015\times\$3/10^6\approx\$0.117,\\
d_{\text{break-even}}
  &=\$2{,}800/(1{,}000\times\$0.117)\\
  &\approx24\ \text{days}\approx3.4\ \text{weeks}.
\end{align*}
This point lies within the 3--6 week scenario range stated in the main paper. Holding the other quantities fixed, an effective billed input-token price of approximately \$1.7 per million gives a 42-day (six-week) break-even, while \$3 per million gives the 24-day point above. Thus the range reflects pricing sensitivity rather than a confidence interval.

The calculation credits input-token reduction only. It excludes summarizer-model calls and tokens, Self-Info scorer compute, output-token pricing, CPU/server cost, retries, latency, integration and maintenance effort, distribution shift, and any monetary value assigned to a change in task success. It therefore should not be read as net profit, annual ROI, a guarantee of permanent savings, or evidence that the policy will not degrade. A deployment must recompute the expression from its billed costs and observed traffic.

\section{Released Data, Code, and Reproduction Boundary}
\label{app:release}

The accompanying \texttt{code\_supplement} contains the reference Beta policy, feature construction, outcome model, event-level training loop, policy server, compressor interfaces, reward and credit functions, statistical utilities, and API-adapter boundary. It also contains protocol scripts for fixed baselines and tool-type grid search, held-out evaluation, single-output interventions, cross-backbone transfer, AGORA-style evaluation, and LOCA-bench evaluation; offline unit tests; pinned Python dependencies; and an anonymized query-schema file.

The query file preserves the nominal 120-query composition, difficulty labels, and 80/40 split, including the 7/19/14 held-out stratification. Some evaluation targets are deliberately redacted or empty. The loaders mark those records unscorable rather than treating them as failures. Thus the file documents schema and split construction but is not a drop-in replacement for the private evaluation set.

The supplement does not include the proprietary production agent, warehouse or knowledge-base contents, raw production prompts and tool outputs, event logs and trajectories, trained checkpoints, or the raw matched inputs from which the paper tables were computed. It also does not vendor LOCA-bench or commercial model endpoints. \texttt{api\_client.py} specifies an integration contract rather than implementing the production backend; evaluation commands require the corresponding external backend, checkpoint, trajectory, or benchmark artifacts and fail when they are absent.

Accordingly, the artifact supports source inspection, offline semantic tests, and reruns against a separately configured compatible backend. It does not support exact end-to-end replay of the production experiments. 

\end{document}